\documentclass[10pt, journal, twoside]{IEEEtran}

\usepackage{amsmath}
\usepackage{graphicx}
\usepackage{amsthm}
\usepackage{amssymb}
\usepackage[ruled]{algorithm2e}
\usepackage{algpseudocode}
\usepackage{microtype}
\usepackage[hidelinks]{hyperref}
\usepackage{xcolor}

\usepackage[caption=false]{subfig}

\DontPrintSemicolon
\SetKwRepeat{doparallel}{do in parallel \ }{while}

\newtheorem{definition}{Definition}

\newcommand{\norm}[1]{\Vert #1 \Vert}

\newcommand{\mc}[1]{\mathcal{#1}}

\newcommand{\ti}[1]{{\tilde{#1}}}

\DeclareMathOperator*{\argmin}{argmin}
\def\*#1{\mathbf{#1}}
\def\+#1{\mathcal{#1}}
\def\<#1{\mathbb{#1}}
\usepackage{adjustbox}
\usepackage{array}

\newcolumntype{R}[2]{%
    >{\adjustbox{angle=#1,lap=\width-(#2)}\bgroup}%
    l%
    <{\egroup}%
}

\begin{document}
\title{Distributed Optimization Methods for Multi-Robot Systems: Part II --- A Survey}

\author{~\IEEEmembership{}Ola~Shorinwa$^1$,~Trevor~Halsted$^1$,~\IEEEmembership{}Javier~Yu$^2$,~\IEEEmembership{}Mac~Schwager$^2$~\IEEEmembership{}%
\thanks{*This project was funded in part by NSF NRI awards 1830402 and 1925030.  The second author was supported on an NDSEG Fellowship, and the third author was supported on an NSF Graduate Research Fellowship.}%
\thanks{$^{**}$ The first three authors contributed equally.}%
\thanks{$^{1}$Department of Mechanical Engineering, Stanford University, Stanford, CA 94305, USA, {\tt\small \{halsted, shorinwa\}@stanford.edu}}%
\thanks{$^{2}$Department of Aeronautics and Astronautics, Stanford University, Stanford, CA 94305, USA
        {\tt\small \{javieryu, schwager\}@stanford.edu}}%
        }
    
\maketitle

\begin{abstract} Although the field of distributed optimization is well-developed, relevant literature focused on the application of distributed optimization to multi-robot problems is limited. This survey constitutes the second part of a two-part series on distributed optimization applied to multi-robot problems. In this paper, we survey three main classes of distributed optimization algorithms---distributed first-order methods, distributed sequential convex programming methods, and alternating direction method of multipliers (ADMM) methods---focusing on fully-distributed methods that do not require coordination or computation by a central computer. We describe the fundamental structure of each category and note important variations around this structure, designed to address its associated drawbacks. Further, we provide practical implications of noteworthy assumptions made by distributed optimization algorithms, noting the classes of robotics problems suitable for these algorithms. Moreover, we identify important open research challenges in distributed optimization, specifically for robotics problems.

\end{abstract}


\section{Introduction}   
In this paper we survey the literature in distributed optimization, specifically with an eye toward problems in multi-robot coordination.  As we demonstrated in the first paper in this two-part series \cite{shorinwa_distributed_2023}, many multi-robot problems can be written as a sum of local objective functions, subject to an intersection of local constraints.  Such problems can be solved with a powerful and growing arsenal of distributed optimization algorithms.  Distributed optimization consists of multiple computation nodes working together to minimize a common objective function through local computation iterations and network-constrained communication steps, providing both computational and communication benefits by eliminating the need for data aggregation.  Distributed optimization is also robust against the failure of individual nodes, as it does not rely on a central computation station, and many distributed optimization algorithms have inherent privacy-preserving properties, keeping the local data, objective function, and constraint function private to each robot, while still allowing for all robots to benefit from one another. Distributed optimization has not yet been widely employed in robotics, and there exist many open opportunities for research in this space, which we highlight in this survey.

Although the field of distributed optimization is well-established in many areas such as computer networking and power systems, problems in robotics have a number of distinguishing features which are not often considered in the major application areas of distributed optimization. Notably, robots move, unlike their analogous counterparts in these other disciplines, which makes their networks time-varying and prone to bandwidth limitations, packet drops, and delays. Robots often use optimization within a receding horizon or model predictive control loop, so fast convergence to an optimal solution is essential in robotics. In addition, optimization problems in robotics are often constrained (e.g., with safety constraints, input constraints, or kino-dynamics constraints in planning problems), and non-convex (for example, simultaneous localization and mapping (SLAM) is a non-convex optimization, as is trajectory planning and state estimation for any nonlinear robot model). Many existing surveys on distributed optimization do not address these unique characteristics of robotics problems.

This survey constitutes the second part of a two-part series on distributed optimization for multi-robot systems.  {\color{black} The first part consists of a tutorial focused on the applicability of distributed optimization to multi-robot problems. In it, we demonstrate how a broad range of multi-robot problems can be cast in a form that is appropriate for distributed optimization, and we provide practical guidelines for implementing distributed optimization algorithms.} In this survey, we highlight relevant distributed optimization algorithms and note the classes of robotics problems to which these algorithms can be applied. Noting the large body of work in distributed optimization, we categorize distributed optimization algorithms into three broad classes and identify the practical implications of these algorithms for robotics problems, including the challenges arising in the implementation of these algorithms on robotics platforms. 

This survey is aimed at robotics researchers, who are interested in research at the intersection of distributed optimization and multi-robot systems, as well as robotics practitioners who want to harness the benefits of distributed optimization algorithms in solving practical robotics problems. In this survey, we limit our discussion to optimization problems over real-valued decision variables.  Although discrete optimization problems (i.e., integer programs or mixed integer programs) arise in some robotics applications, these problems are beyond the scope of this survey.  However, we note that distributed algorithms for integer and mixed integer problems have been discussed in a number of different works \cite{prodan2014mixed, murray2018hierarchical, testa2019distributed}. Further, we limit our discussion to derivative-based methods, in contrast to derivative-free (zeroth-order) distributed optimization algorithms. We note that derivative-free optimization methods have been discussed extensively in \cite{liu2020primer, hajinezhad2017zeroth, hajinezhad2019zone, hajinezhad2019perturbed, beznosikov2020derivative, tang2020distributed}.

In many robotics applications, such as field robotics, communication with a central computer (or the cloud) might be infeasible, even though each robot can communicate locally with other neighboring robots. Consequently, we focus particularly on distributed optimization algorithms that permit robots to use local robot-to-robot communication to compute an optimal solution, rather than algorithms that require coordination by a central computer. These methods yield a globally optimal solution for convex problems and, in general, a locally optimal solution for non-convex problems, producing the same quality solution that would be obtained if a centralized method were applied. {\color{black} Although many distributed optimization algorithms are not inherently ``online" (in the sense that these algorithms were not originally designed to be executed while the robot is actively gathering data or completing a task, providing information that changes its objective and constraint functions), we note that many of these algorithms can be applied in these online problems within the model predictive control (MPC) framework, where a new optimization problem is solved periodically from streaming data.} %

\begin{figure}[t!]
\centering
\subfloat[Optimization by one robot yields the solution given only that robot's observations.]{%
  \includegraphics[width=0.8\columnwidth]{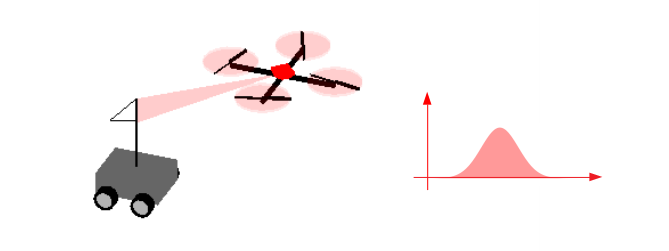}%
}

\subfloat[Using distributed optimization, each robot obtains the optimal solution resulting from all robots' observations.]{%
  \includegraphics[width=0.8\columnwidth]{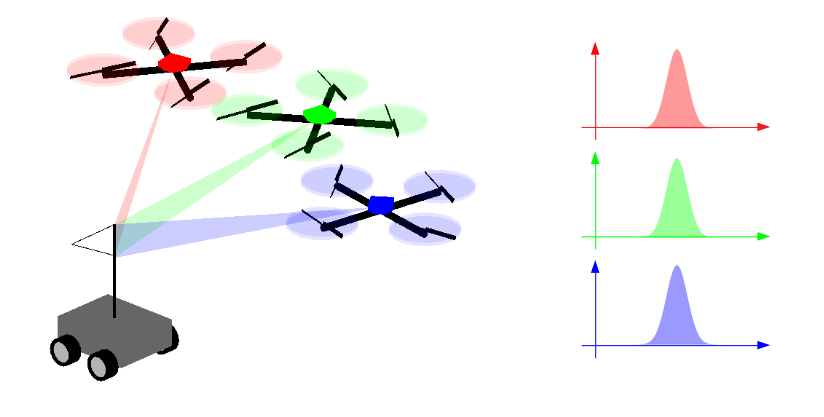}%
  
}

\caption{A motivation for distributed optimization: consider an estimation scenario in which a robot seeks to localize a target given sensor measurements. The robot can compute an optimal solution given \emph{only its observations}, as represented in (a). By using distributed optimization techniques, each robot in a networked system of robots can compute the optimal solution \emph{given all robots' observations} without actually sharing individual sensor models or measurements with one another, as represented in (b).}

\end{figure}

In this survey, we provide a taxonomy of the different algorithms for performing distributed optimization based on their defining mathematical characteristics.  We identify three classes: distributed first-order algorithms, distributed sequential convex programming, and distributed extensions to the alternating direction method of multipliers (ADMM).

\textbf{Distributed First-Order Algorithms:} The most common class of distributed optimization methods is based on the idea of averaging local gradients computed by each computational node to perform an approximate gradient descent update \cite{tsitsiklis1986}, and in this work, we refer to them as Distributed First-Order (DFO) algorithms. DFO algorithms can be further sub-divided into distributed (sub)-gradient descent, distributed gradient tracking, distributed stochastic gradient descent, and distributed dual averaging algorithms, with each sub-category differing from the others based on the order of the update steps and the nature of the gradients used. In general, DFO algorithms use consensus methods to achieve a shared solution for the optimization problem. Many DFO algorithms allow for dynamic communication networks (including uni-directional and bi-directional networks) \cite{saadatniaki2018optimization, xin2018linear} and limited computation resources \cite{nedic2009}, but they are often not well-suited to constrained problems.

\textbf{Distributed Sequential Convex Programming:} Sequential Convex Optimization is a common technique in centralized optimization that involves minimizing a sequence of convex approximations to the original (usually non-convex) problem. Under certain conditions, the sequence of sub-problems converges to a local optimum of the original problem. Newton's method and the 
Broyden–Fletcher–Goldfarb–Shanno (BFGS) method are common examples. The same concepts are used by a number of distributed optimization algorithms, and we refer to these algorithms as Distributed Sequential Convex Programming methods. Generally, these methods use consensus techniques to construct the convex approximations of the joint objective function. One example is the Network Newton method \cite{Mokhtari2015}, which uses consensus to approximate the inverse Hessian of the objective to construct a quadratic approximation of the joint problem. The NEXT family of algorithms \cite{di2016next} provides a flexible framework, which can utilize a variety of convex surrogate functions to approximate the joint problem, and is specifically designed to optimize non-convex objective functions. Although many distributed sequential convex programming methods are not suitable for problems with dynamic communication networks, a few distributed sequential convex programming algorithms are amenable to these problems \cite{di2016next}.

\textbf{Alternating Direction Method of Multipliers:} The last class of algorithms covered in this survey is based on the alternating direction method of multipliers (ADMM) \cite{rockafellar1976monotone}. ADMM works by minimizing the augmented Lagrangian of the optimization problem using alternating updates to the primal and dual variables \cite{boyd2011distributed}. {\color{black} This method naturally accommodates constrained problems  (with the assumption that we can convert inequality constraints to equality constraints using slack variables)}. The original method is distributed, but not in the sense we consider in this survey. Specifically, the original ADMM requires a central computation hub to collect all local primal computations from the nodes to perform a centralized dual update step. ADMM was first modified to remove this requirement for a central node in \cite{mateos2010distributed}, where it was used for distributed signal processing. The algorithm from \cite{mateos2010distributed} has since become known as Consensus ADMM (C-ADMM), although the original paper \cite{mateos2010distributed} did not use this terminology. A number of other distributed variants have been developed to address many unique characteristics, including uni-directional communication networks and limited communication bandwidth \cite{khatana2020dc, zhu2016quantized}, which are often present in robotics problems.

    \subsection{Existing Surveys}
    A number of other recent surveys on distributed optimization exist, and provide useful background when working with the algorithms covered in this survey. Some of these surveys cover applications of distributed optimization in distributed power systems \cite{molzahn2017survey}, big-data problems \cite{scutari2018parallel}, and game theory \cite{yang2010distributed}, while others focus primarily on first-order methods for problems in multi-agent control \cite{nedic2018distributed}. Other articles broadly address distributed first-order optimization methods, including a discussion on the communication-computation trade-offs \cite{nedic2018network, nedic2018distributedcollection}. Another survey \cite{chang2020distributed} covers exclusively non-convex optimization in both batch and data-streaming contexts, but again only analyzes first-order methods. Finally, \cite{yang2019survey} covers a wide breadth of distributed optimization algorithms with a variety of assumptions, focusing exclusively on convex optimization problems. {\color{black}Building on the first paper in the series \cite{shorinwa_distributed_2023} which
    formulates multi-robot problems within the framework of distributed optimization, our survey differs from other existing surveys in that it specifically targets applications of distributed optimization to multi-robot problems: identifying suitable distributed optimization algorithms that address the practical issues arising in multi-robot problems and providing references demonstrating the application of distributed optimization to multi-robot problems.} As a result, this survey highlights the practical implications of the assumptions made by many distributed optimization algorithms and provides a condensed taxonomic overview of useful methods for these applications. Other useful background material can be found for distributed computation \cite{bertsekas1989parallel} \cite{lynch1996distributed}, and on multi-robot systems in \cite{distctrlrobotnetw} \cite{mesbahi2010graph}. 
    
    \subsection{Contributions}
    This survey paper has three primary objectives:
    \begin{enumerate}
        \item Survey the literature across three different classes of distributed optimization algorithms, noting the defining mathematical characteristics of each category.
        \item Highlight noteworthy assumptions made by distributed optimization algorithms, and provide existing applications of distributed optimization algorithms to multi-robot problems.
        \item Propose open research problems in distributed optimization for robotics.
    \end{enumerate}
    
    \subsection{Organization}
    In Section \ref{sec:preliminaries} we introduce mathematical notation and preliminaries, and in Section  \ref{sec:problem_formulation} we present the general formulation for the distributed optimization problem and describe the general framework shared by distributed optimization algorithms. Sections {\ref{sec:DistGradDesc}--\ref{sec:ADMM}} survey the literature in each of the three categories, and provide details for representative algorithms in each category. Section \ref{sec:applications_in_literature} provides existing applications of distributed optimization in the robotics literature.
    In Section~\ref{sec:open_problems}, we discuss open research problems in applying distributed optimization to multi-robot systems and robotics in general, and we offer concluding remarks in Section~\ref{sec:conclusion}.

    \section{Notation and Preliminaries}
    \label{sec:preliminaries}
    In this section, we introduce the notation used in this paper and provide the definitions of mathematical concepts relevant to the discussion of the distribution optimization algorithms. We denote the gradient of a function ${f: \mathbb{R}^{n} \rightarrow \mathbb{R}}$ as ${\nabla f}$ and its Hessian as ${\nabla^{2} f}$. We denote the vector containing all ones as $\mathbf{1}_{n}$, where $n$ represents the number of elements in the vector. Next, we begin with the definition of stochastic matrices which arise in distributed first-order optimization algorithms.
    
    \begin{definition}[Non-negative Matrix]
        A matrix ${W \in \mathbb{R}^{n \times n}}$ is referred to as a non-negative matrix if ${w_{ij} \geq 0}$ for all ${i, j \in \{1,\cdots,n\}}$.
    \end{definition}
    
    \begin{definition}[Stochastic Matrix]
        A non-negative matrix ${W \in \mathbb{R}^{n \times n}}$ is referred to as a row-stochastic matrix if
        \begin{equation}
            W \mathbf{1}_{n} = \mathbf{1}_{n},
        \end{equation}
        in other words, the sum of all elements in each row of the matrix equals one. 
        We refer to $W$ as a column-stochastic matrix if
        \begin{equation}
            \mathbf{1}_{n}^{\top}W = \mathbf{1}^\top_{n}.
        \end{equation}
        Likewise, for a doubly-stochastic matrix $W$,
        \begin{equation}
            W \mathbf{1}_{n} = \mathbf{1}_{n}\ \text{and}\ \mathbf{1}_{n}^{\top}W = \mathbf{1}^\top_{n}.
        \end{equation}
    \end{definition}
    
    Now, we provide the definition of some relevant properties of a sequence.
    
    \begin{definition}[Summable Sequence]
        A sequence $\{\alpha(k)\}_{k \geq 0}$, with ${k \in \mathbb{N}}$, is a summable sequence {\color{black}{if $\alpha(k) > 0$ for all $k$ and}}
        \begin{equation}
            \sum_{k = 0}^{\infty} \alpha(k) < \infty.
        \end{equation}
    \end{definition}
    
    \begin{definition}[Square-Summable Sequence]
        A sequence $\{\alpha(k)\}_{k \geq 0}$, with ${k \in \mathbb{N}}$, is a square-summable sequence {\color{black}{if $\alpha(k) > 0$ for all $k$ and}}
        \begin{equation}
            \sum_{k = 0}^{\infty} \left(\alpha(k)\right)^{2} < \infty.
        \end{equation}
    \end{definition}
    
    We discuss some relevant notions of the connectivity of a graph.
    
    \begin{definition}[Connectivity of an Undirected Graph]
        An undirected graph $\mathcal{G}$ is connected if a path exists between every pair of vertices $(i,j)$ where ${i,j \in \mathcal{V}}$. Note that such a path might traverse other vertices in $\mathcal{G}$.
    \end{definition}
    
    \begin{definition}[Connectivity of a Directed Graph]
        A directed graph $\mathcal{G}$ is strongly connected if a directed path exists between every pair of vertices $(i,j)$ where ${i,j \in \mathcal{V}}$. In addition, a directed graph $\mathcal{G}$ is weakly connected if the underlying undirected graph is connected. The underlying undirected graph $\mathcal{G}_{u}$ of a directed graph $\mathcal{G}$ refers to a graph with the same set of vertices as $\mathcal{G}$ and a set of edges obtained by considering each edge in $\mathcal{G}$ as a bi-directional edge. Consequently, every strongly connected directed graph is weakly connected; however, the converse is not true.
    \end{definition}

	In distributed optimization in multi-robot systems, robots perform communication and computation steps to minimize some global objective function. We focus on problems in which the robots' exchange of information must respect the topology of an underlying distributed communication graph, which could possibly change over time.  This communication graph, denoted as $\mc{G}(t) = (\mc{V}(t), \mc{E}(t))$, consists of vertices $\mc{V}(t) = \{1, \dots, N\}$ and edges $\mc{E}(t) \subseteq \mc{V}(t) \times \mc{V}(t)$ over which pairwise communication can occur. For undirected graphs, we denote the set of neighbors of robot $i$ as $\mathcal{N}_{i}(t)$. For directed graphs, we refer to the set of robots which can \textit{send} information to robot $i$ as the set of in-neighbors of robot $i$, denoted by $\mathcal{N}_{i}^{+}(t)$. Likewise, for directed graphs, we refer to the set of robots which can \textit{receive} information from robot $i$ as the out-neighbors of robot $i$, denoted by $\mathcal{N}_{i}^{-}(t)$.
    
    \begin{definition}[Convergence Rate]
        Provided that a sequence $\{x^{(k)}\}$ converges to $x^{\star}$, if there exists a positive scalar ${r \in \mathbb{R}}$, with ${r \geq 1}$, and a constant ${\lambda \in \mathbb{R}}$, with ${\lambda > 0}$, such that
        \begin{equation}
            \lim_{k \rightarrow \infty} \frac{\norm{x^{(k + 1)} - x^{\star}}}{\norm{x^{(k)} - x^{\star}}^{r}} = \lambda,
        \end{equation}
        then $r$ defines the order of convergence of the sequence $\{x^{(k)}\}$ to $x^{\star}$. Moreover, the asymptotic error constant is given by $\lambda$.
        
        If ${r = 1}$ and ${\lambda = 1}$, then $\{x^{(k)}\}$ converges to $x^{\star}$ sub-linearly. However, if ${r = 1}$ and ${\lambda < 1}$, then $\{x^{(k)}\}$ converges to $x^{\star}$ linearly. Likewise, $\{x^{(k)}\}$ converges to $x^{\star}$ quadratically if ${r = 2}$ and cubically if ${r = 3}$.
    \end{definition}
    
    \begin{definition}[Synchronous Algorithm]
        \label{def:sync}
        An algorithm is synchronous if each robot (computational node) has to wait at a predetermined point for a specific message from other robots (computational nodes) before proceeding. In general, the end of an iteration of the algorithm represents the predetermined synchronization point. Conversely, in an asynchronous algorithm, each robot completes each iteration at its own pace, without having to wait at a predetermined point. In other words, at any given time, the number of iterations of an asynchronous algorithm completed by each robot could differ from the number of iterations completed by other robots.
    \end{definition}

	\section{Problem Formulation}
	\label{sec:problem_formulation}

    {\color{black} We consider a general class of \emph{separable} distributed optimization problems, in which we express a \emph{joint} objective function as the sum over \emph{local} objective functions.  From a multi-robot perspective, each robot only knows its own local function, but the robots collectively seek to find the optimum to the global function. In this general formulation, we also consider a set of joint constraints consisting of an intersection over local constraints.  Each robot only knows its own local constraints and its local objective function. The resulting optimization problem is given by}
	\begin{align}
    	\label{eq:general_problem}
    	\begin{split}
    	\min_x \:&\sum_{i \in \mc{V}} f_i(x)\\
    	\text{subject to }&g_i(x) = 0 \quad \forall i \in \mc{V}
    	\\& h_i(x) \le 0 \quad \forall i \in \mc{V}
    	\end{split}
	\end{align}
	where ${x \in \mathbb{R}^{n}}$ denotes the optimization variable and ${f_{i}: \mathbb{R}^{n} \rightarrow \mathbb{R}}$, ${g_{i}: \mathbb{R}^{n} \rightarrow \mathbb{R}}$, and ${h_{i}: \mathbb{R}^{n} \rightarrow \mathbb{R}}$ denote the local objective function, equality constraint function, and inequality constraint function of robot $i$, respectively. The joint optimization problem \eqref{eq:general_problem} can be solved locally by each robot if all the robots share their objective and constraint functions with one another. Alternatively, the solution can be computed centrally if all the local functions are collated at a central station. However, robots typically possess limited computation and communication resources, which precludes each robot from sharing its local functions with other robots, particularly in problems with high-dimensional problem data, such as images, lidar and other perception measurements. 
 
    Distributed optimization algorithms enable each robot to compute a solution of \eqref{eq:general_problem} locally without sharing its local objective, constraints, or data. These algorithms assign a copy of the optimization variable to each robot, enabling each robot to update its own copy locally and in parallel with other robots. Moreover, distributed optimization algorithms enforce consensus among the robots for agreement on a common solution of the optimization problem. Consequently, these algorithms solve an equivalent reformulation of the optimization problem in \eqref{eq:general_problem}, given by
    \begin{align}
    	\label{eq:general_problem_with_agreement_explicit}
    	\begin{split}
    	\min_{\{x_i,\ \forall i \in \mc{V}\}} \:&\sum_{i \in \mc{V}} f_i(x_i)\\
    	\text{subject to }& x_i = x_j \quad \forall (i, j) \in \mc{E} \\&g_i(x_i) = 0 \quad \forall i \in \mc{V}
    	\\& h_i(x_i) \le 0 \quad \forall i \in \mc{V},
    	\end{split}
	\end{align}
	where ${x_{i} \in \mathbb{R}^{n}}$ denotes robot $i$'s local copy of the optimization variable. We note that the consensus constraints in \eqref{eq:general_problem_with_agreement_explicit} ensure agreement among all the robots, with the assumption that the communication graph is connected. Moreover, the consensus constraints are enforced between neighboring robots only, making it compatible with a point-to-point communication network, where robots can only communicate with their one-hop neighbors. To simplify notation, we introduce the set ${\mathcal{X}_{i} = \{x_{i} \mid g_i(x_i) = 0, h_i(x_i) \le 0\}}$, representing the feasible set given the constraint functions $g_{i}$ and $h_{i}$. Consequently, we can express the problem in \eqref{eq:general_problem_with_agreement_explicit} succinctly as follows:
    \begin{align}
    	\label{eq:general_problem_with_agreement}
    	\begin{split}
    	\min_{\{x_i \in \mathcal{X}_{i},\ \forall i \in \mc{V}\}} \:&\sum_{i \in \mc{V}} f_i(x_i)\\
    	\text{subject to }& x_i = x_j \quad \forall (i, j) \in \mc{E}.
    	\end{split}
    \end{align}
	
	In the following sections, we discuss three broad classes of distributed optimization methods, namely, distributed first-order methods, distributed sequential convex programming methods, and the alternating direction method of multipliers. We note that distributed first-order methods and distributed sequential convex programming methods implicitly enforce the consensus constraints in \eqref{eq:general_problem_with_agreement}, while the alternating direction method of multipliers enforces these constraints explicitly.
    {\color{black} While not all of the methods that we survey explicitly address constraints of the form $g_i(x) = 0$, $h_i(x) \le 0$, we note in each section considerations to accommodate these additional terms. In some cases, it is also appropriate to incorporate the constraints as penalty terms in the cost function.}
	
	Before proceeding, we highlight the general framework that distributed optimization algorithms share. Distributed optimization algorithms are iterative algorithms in which each robot executes a number of operations over discrete iterations $k = 0, 1, \dots$ until convergence, where each iteration consists of a communication and computation step. During each communication round, each robot shares a set of its local variables with its neighbors, referred to as its ``communicated'' variables $\mc{Q}_i^{(k)}$, which we distinguish from its ``internal'' variables $\mc{P}_i^{(k)}$, which are not shared with its neighbors. In general, each algorithm requires initialization of the local variables of each robot, in addition to algorithm-specific parameters, denoted by  $\mc{R}_i^{(k)}$. {\color{black} We note that some algorithms require all the robots to utilize a common step-size at initialization; however, these parameters can be initialized prior to deployment of the robots.}

	\section{\texorpdfstring{Distributed First-Order Algorithms}{Distributed First-Order Algorithms}}
	\label{sec:DistGradDesc}
	The optimization problem in \eqref{eq:general_problem} (in its unconstrained form) can be solved through gradient descent where the optimization variable is updated using
	\begin{equation}
	    x^{(k+1)} = x^{(k)} - \alpha^{(k)} \nabla f(x^{(k)}) \label{eqn:cent_grad_desc}
	\end{equation}
	with $\nabla f(x^{(k)})$ denoting the gradient of the objective function at $x^{(k)}$, given by
	\begin{equation}
	    \nabla f(x) = \sum _{i \in \mathcal{V}} \nabla f_i(x),
	\end{equation}
	given some scheduled step-size $\alpha^{(k)}$. Inherently, computation of $\nabla f(x^{(k)})$ requires knowledge of the local objective functions or gradients by all robots in the network which is infeasible in many problems.
	
	Distributed First-Order (DFO) algorithms extend the centralized gradient scheme to the distributed setting where robots communicate with one-hop neighbors without knowledge of the local objective functions or gradients of all robots. In DFO methods, each robot updates its local variable using a weighted combination of the local variables or gradients of its neighbors according to the weights specified by a stochastic weighting matrix $W$, allowing for the dispersion of information on the objective function or its gradient through the network. The stochastic matrix $W$ must be compatible with the underlying communication network, with a non-zero element $w_{ij}$ when robot $j$ can send information to robot $i$.

 {\color{black} From the perspective of a single robot, the update equations in DFO methods
represent a trade-off between optimality of its individual solution based on its local
objective function and agreement with its neighbors. Consensus enables the robot to incorporate global information about the objective function's shape into its update, thereby allowing it to approximate a gradient descent step on the global cost function rather than on its local cost function.}
	
	Many DFO algorithms use a doubly-stochastic matrix, a row-stochastic matrix \cite{mai2019distributed}, or a column-stochastic matrix, depending on the model of the communication network considered, while other methods use a push-sum approach. In addition, many methods further require symmetry of the doubly-stochastic weighting matrix with ${W = W^{\top}}$. The weight matrix exerts a significant influence on the convergence rates of DFO algorithms, and thus, an appropriate choice of these weights are required for convergence of DFO methods.
	
	The order of the update procedures for the local variables of each robot and the gradient used by each robot in performing its local update procedures differ among DFO algorithms, giving rise to four broad classes of DFO methods: Distributed (Sub)-Gradient Descent and Diffusion Algorithms, Gradient Tracking Algorithms, Distributed Stochastic Gradient Algorithms, and Distributed Dual Averaging. While distributed (sub)-gradient descent algorithms require a decreasing step-size for convergence to an optimal solution, gradient tracking algorithms converge to an optimal solution without this condition. We discuss these distributed methods in the following subsections.

	\subsection{Distributed (Sub)-Gradient Descent and Diffusion Algorithms}

    \begin{algorithm}[t]
    	\caption{Distributed Gradient Descent (DGD)}\label{alg:DistGradDesc}
    	\textbf{Initialization:} $k \gets 0$, ${x_{i}^{(0)} \in \mathbb{R}^{n}}$
    	\;
        \textbf{Internal variables:} $\mc{P}_i^{(k)} = \emptyset$
        \;
        \textbf{Communicated variables:} $\mc{Q}_i^{(k)} = x_i^{(k)}$
        \;
        \textbf{Parameters:} $\mc{R}_i^{(k)} = (\alpha^{(k)}, w_i)$\;
        
        \doparallel ($\forall i \in \mc{V}$){stopping criterion is not satisfied} {
    	    Communicate $\mc{Q}_i^{(k)}$ to all $j \in \mc{N}_i$\;
    	    Receive $\mc{Q}_j^{(k)}$ from all $j \in \mc{N}_i$
    	    \begin{flalign*}
                \alpha^{(k)} &= \frac{\alpha^{(0)}}{\sqrt{k}} &\\
        	    x_i^{(k+1)} &= \sum_{j \in \mathcal{N}_i \cup \{i\}} w_{ij} x_j^{(k)} - \alpha^{(k)} \nabla f_i(x_i^{(k)}) &
    	    \end{flalign*}
    	    
    	    $k \gets k + 1$
        }
    \end{algorithm}

	Tsitsiklis introduced a model for distributed gradient descent in the 1980s in \cite{tsitsiklis1984problems} and \cite{tsitsiklis1986} (see also \cite{bertsekas1989parallel}). The works of Nedi\'{c} and Ozdaglar in \cite{nedic2009} revisit the problem, marking the beginning of interest in consensus-based frameworks for distributed optimization over the recent decade. This basic model of distributed gradient descent consists of an update term that involves consensus on the optimization variable as well as a step in the direction of the local gradient for each node:
	\begin{equation}
	    x_i^{(k+1)} = \sum_{j \in \mathcal{N}_i \cup \{i\}} w_{ij} x_j^{(k)} - \alpha_i^{(k)} \nabla f_i\left(x_i^{(k)}\right) \label{eqn:basic_DGD}
	\end{equation}
	where robot $i$ updates its variable using a weighted combination of its neighbors' variables determined by the weights $w_{ij}$ with $\alpha_{i}(k)$ denoting its local step-size at iteration $k$. 
	
	For convergence to the optimal joint solution, these methods require the step-size to asymptotically decay to zero. As proven in \cite{lobel2010distributed}, if $\alpha^{(k)}$ is chosen such that the sequence $\{\alpha^{(k)}\}$ is square-summable but not summable, then the optimization variables of all robots converge to the optimal joint solution, given the standard assumptions of a connected network, properly chosen weights, and bounded (sub)-gradients. In contrast, the choice of a constant step-size for all time-steps only guarantees convergence of each robot's iterates to a neighborhood of the optimal joint solution.
 {\color{black} In practice, this means that a multi-robot system implementing distributed gradient descent must coordinate on scheduling the decrease of the step size. Nonetheless, distributed gradient descent can generally tolerate some level of asynchrony or stochasticity.}
    Algorithm \ref{alg:DistGradDesc} summarizes the update step for the distributed gradient descent method in \cite{nedic2009} with the step-size ${\alpha^{(k)} = \frac{\alpha^{(0)}}{\sqrt{k}}}$, with ${\alpha^{(0)} > 0}$.
    Notably, this step-size is neither summable nor square-summable. In fact, only the non-summable condition on the step-size is necessary for convergence to the optimal joint solution.

	We note that the update procedure given in \eqref{eqn:basic_DGD} requires a doubly-stochastic weighting matrix, which, in general, is incompatible with directed communication networks. Other distributed gradient descent algorithms \cite{olshevsky2009convergence, olshevsky2018robust, nedic2014distributed, benezit2010weighted} utilize the \textit{push-sum} consensus protocol \cite{kempe2003gossip} in place of the consensus terms in \eqref{eqn:basic_DGD}, extending the application of distributed gradient descent schemes to problems with directed communication networks.

	In general, with a constant step-size, distributed (sub)-gradient descent algorithms converge at a rate of ${O(1 / k)}$ to a neighborhood of the optimal solution in convex problems \cite{yuan2016convergence}. With a decreasing step-size, some distributed (sub)-gradient descent algorithms converge to an optimal solution at $O(\log k / k)$ using an accelerated gradient scheme such as the Nesterov gradient method \cite{jakovetic2014fast}.
	
    \subsection{Distributed Gradient Tracking Algorithms}

    \begin{algorithm}[t]
    	\caption{DIGing}\label{alg:DIGing}
    	\textbf{Initialization:} $k \gets 0$, ${x_{i}^{(0)} \in \mathbb{R}^{n}}$, ${y_{i}^{(0)} = \nabla f_{i}(x_{i}^{(0)})}$
    	\;
        \textbf{Internal variables:} $\mc{P}_i^{(k)} = \emptyset$
        \;
        \textbf{Communicated variables:} $\mc{Q}_i^{(k)} = \left(x_i^{(k)}, y_i^{k}\right)$
        \;
        \textbf{Parameters:} $\mc{R}_i^{(k)} = \left(\alpha, w_i\right)$\;
        
        \doparallel ($\forall i \in \mc{V}$){stopping criterion is not satisfied} {
    	    Communicate $\mc{Q}_i^{(k)}$ to all $j \in \mc{N}_i$\;
    	    Receive $\mc{Q}_j^{(k)}$ from all $j \in \mc{N}_i$
    	    \begin{flalign*}
        	    x_i^{(k+1)} &= \sum_{j \in \mc{N}_i \cup \{i\}} w_{ij}x_j^{(k)} - \alpha y_i^{(k)} &\\
        	    y_i^{(k+1)} &= \sum_{j \in \mc{N}_i \cup \{i\}} w_{ij} y_j^{(k)} + \nabla f_i(x_i^{(k+1)}) - \nabla f_i(x_i^{(k)}) &
    	    \end{flalign*}
    	    
    	    $k \gets k + 1$
        }
    \end{algorithm}

	Although distributed (sub)-gradient descent algorithms converge to an optimal joint solution, the requirement of a square-summable sequence $\{\alpha^{(k)}\}$ --- which results in a decaying step-size --- reduces the convergence speed of these methods. Gradient tracking methods address this limitation by allowing each robot to utilize the changes in its local gradient between successive iterations as well as a local estimate of the average gradient across all robots in its update procedures, enabling the use of a constant step-size while retaining convergence to the optimal joint solution.
	
	The EXTRA algorithm introduced by Shi \textit{et al.} in \cite{shi2015extra} uses a fixed step-size while still achieving exact convergence. EXTRA replaces the gradient term with the difference in the gradients of the previous two iterates. Because the contribution of this gradient difference term decays as the iterates converge to the optimal joint solution, EXTRA does not require the step-size to decay in order to converge to the exact optimal joint solution. EXTRA achieves linear convergence \cite{yuan2016convergence}, and a variety of gradient tracking algorithms have since offered improvements on its linear rate \cite{daneshmand2018second}, for convex problems with strongly convex objective functions.
	
	The DIGing algorithm \cite{nedic2017achieving, nedic2017digging}, whose update equations are shown in Algorithm \ref{alg:DIGing}, is one such similar method that extends the faster convergence properties of EXTRA to the domain of directed and time-varying graphs. DIGing requires communication of two variables, effectively doubling the communication cost per iteration when compared to DGD, but greatly increasing the diversity of communication infrastructure that it can be deployed on.

	Many other gradient tracking algorithms involve variations on the variables updated using consensus and the order of the update steps, such as NIDS \cite{li2019decentralized}, Exact Diffusion \cite{yuan2018exact, yuan2018exact2}, \cite{qu2017harnessing}, and \cite{xin2019distributed}. These algorithms, which generally require the use of doubly-stochastic weighting matrices, have been extended to problems with row-stochastic or column-stochastic matrices \cite{saadatniaki2018optimization, xi2017add, xin2018linear, xi2018linear} and push-sum consensus \cite{zeng2017extrapush} for distributed optimization in directed networks. To achieve faster convergence rates, many of these algorithms require each robot to communicate multiple local variables to its neighbors during each communication round. In addition, we note that some of these algorithms require all robots to use the same step-size, which can prove challenging in some situations. Several works offer a synthesis of various gradient tracking methods, noting the similarities between these methods. Under the canonical form proposed in \cite{sundararajan2019canonical, sundararajan2020analysis}, these algorithms and others differ only in the choice of several constant parameters. 
	Jakoveti\'{c} also provides a unified form for various gradient tracking algorithms in \cite{jakovetic2018unification}. Some other works consider accelerated variants using Nesterov gradient descent \cite{qu2019accelerated, xin2019distributedNesterov, qu2019accelerated, lu2020nesterov}. Gradient tracking algorithms can be considered to be primal-dual methods with an appropriately defined augmented Lagrangian function \cite{nedic2017achieving, mansoori2019general}.
	
    In general, gradient tracking algorithms address unconstrained distributed convex optimization problems, but these methods have been extended to non-convex problems \cite{tatarenko2017non} and constrained problems using projected gradient descent \cite{ram2010distributed, bianchi2012convergence, johansson2010randomized}. Some other methods \cite{maros2018panda, maros2020geometrically, maros2019eco, seaman2017optimal} perform dual-ascent on the dual problem of \eqref{eq:general_problem}, where the robots compute their local primal variables from the related minimization problem using their dual variables. These methods require doubly-stochastic weighting matrices but allow for time-varying communication networks. Distributed first-order methods have been extended to the constrained setting \cite{lan2020communication}, where each robot performs a subsequent proximal projection step to obtain solutions which satisfy the problem constraints. Distributed first-order methods have been further extended to the conjugate gradient setting \cite{shorinwa2024conjugate}, where each robot updates its local variables using its local estimate of the conjugate gradient.

    {\color{black} In deep learning problems, the associated objective function often consists of a sum over a very large number of data points. Computing exact gradients for such problems can be prohibitively costly, so gradients are approximated by randomly sampling a subset of the data at each iteration and computing the gradient only over those data. Such methods, called stochastic gradient descent, dominate in deep learning.}  In \cite{lian2017can}, stochastic gradients are used in place of gradients in the DGD algorithm, and the resulting algorithm is shown to converge.

	\subsection{Distributed Dual Averaging}

    \begin{algorithm}[t]
    	\caption{Distributed Dual Averaging (DDA)}\label{alg:DistDualAvg}
    	\textbf{Initialization:} $k \gets 0$, ${x_{i}^{(0)} \in \mathbb{R}^{n}}$, ${z_{i}^{(0)} = x_{i}^{(0)}}$
    	\;
        \textbf{Internal variables:} $\mc{P}_i = z_i^{(k)}$
        \;
        \textbf{Communicated variables:} $\mc{Q}_i^{(k)} = x_i^{(k)}$
        \;
        \textbf{Parameters:}$\mc{R}_i^{(k)} = \left(\alpha^{(k)}, w_i, \phi(\cdot)\right)$\;
        
        \doparallel ($\forall i \in \mc{V}$){stopping criterion is not satisfied} {
    	    Communicate $\mc{Q}_i^{(k)}$ to all $j \in \mc{N}_i$\;
    	    Receive $\mc{Q}_j^{(k)}$ from all $j \in \mc{N}_i$
    	    \begin{flalign*}
        	    z_i^{(k+1)} &= \sum_{j \in \mc{N}_i \cup \{i\}} w_{ij} z_j^{(k)} + \nabla f_i\left(x_i^{(k)}\right) &\\
        	    x_i^{(k+1)} &= \argmin_{x  \in \mathcal{X}_{i}} \left\{x^\top z_i^{(k+1)} + \frac{1}{\alpha^{(k)}} \phi(x) \right\} &
    	    \end{flalign*}
    	    
    	    $k \gets k + 1$
        }
    \end{algorithm}

    Dual averaging first posed in \cite{nesterov2009primal}, and extended in \cite{xiao2010dual}, takes a similar approach to distributed (sub)-gradient descent methods in solving the optimization problem in \eqref{eq:general_problem}, with the added benefit of providing a mechanism for handling problem constraints through a projection step, in like manner as projected (sub)-gradient descent methods. However, the original formulations of the dual averaging method requires knowledge of all components of the objective function or its gradient which is unavailable to all robots. The Distributed Dual Averaging method (DDA) circumvents this limitation by modifying the update equations using a doubly-stochastic weighting matrix to allow for updates of each robot's variable using its local gradients and a weighted combination of the variables of its neighbors \cite{duchi2011dual}.
    
    Similar to distributed (sub)-gradient descent methods, distributed dual averaging requires a sequence of decreasing step-sizes to converge to the optimal solution. Algorithm \ref{alg:DistDualAvg} provides the update equations in the DDA algorithm, along with the projection step which involves a proximal function $\phi(x)$, often defined as $\frac{1}{2}\norm{x}_2^2$. After the projection step, the robot's variable satisfies the problem constraints described by the constraints set $\mc{X}$. Some of the same extensions made to distributed (sub)-gradient descent algorithms have been studied for DDA, including analysis of the algorithm under communication time delays \cite{tsianos2011distributed} and replacement of the doubly-stochastic weighting matrix with push-sum consensus \cite{tsianos2012push}.

	\section{Distributed Sequential Convex Programming}\label{sec:DistSeqCvx}
	Sequential Convex Programming is a class of optimization methods, typically for non-convex problems, that proceed iteratively by approximating the nonconvex problem with a convex surrogate computed from the current values of the decision variables.  This convex surrogate is optimized, and the resulting decision variables are used to compute the convex surrogate for the next iterate.  Newton's method is a classic example of a Sequential Convex Method, in which the convex surrogate is a quadratic approximation of the original objective function.  Several methods have been proposed for distributed Sequential Convex Programming, as we survey here. 
 {\color{black} As with distributed first-order methods, distributed sequential convex programming takes the perspective of using consensus to approximate the global objective function, with the addition of approximating not only the global gradient but also the global Hessian. The benefit of this approach is that convergence typically requires fewer iterations and is less dependent on carefully selecting a step size.  This comes at the expense of requiring the robots to communicate more information in order to approximate the second-order characteristics of the global objective function. }
	\subsection{Approximate Newton Methods}
    Newton's method and its variants are commonly used for solving convex optimization problems, and they provide significant improvements in convergence rate when second-order function information is available \cite{boyd2004convex}.
    To apply Newton's method to the distributed optimization problem in \eqref{eq:general_problem}, the Network Newton-$K$ (NN-$K$) algorithm \cite{Mokhtari2015} takes a penalty-based approach which introduces consensus between the robots' variables as components of the objective function. %
    The NN-$K$ method reformulates the constrained form of the distributed problem in \eqref{eq:general_problem} as the following unconstrained optimization problem:  
    \begin{equation}
        \label{eq:penalty_distributed_problem}
	    \min_{\{x_{i} \in \mathbb{R}^{n},\ \forall i \in \mathcal{V}\}} \: \alpha \sum_{i \in \mc{V}} f_i(x_{i}) + 
	    x_i^{\top} \left( \sum_{j \in \mc{N} \cup \{i\}} \bar{w}_{ij}x_{j}\right)
    \end{equation}
    where $\bar{W} = I - W$, and $\alpha$ is a weighting hyperparameter.

    \begin{algorithm}[t]
    	\caption{Network Newton-$K$ (NN-$K$)}\label{alg:NetNewtonPa}
    	\textbf{Initialization:} $k \gets 0$, ${x_{i}^{(0)} \in \mathbb{R}^{n}}$
    	\;
        \textbf{Internal variables:} $\mc{P}_i^{(k)} = \left(g_i^{(k)}, D_i^{(k)}\right)$
        \;
        \textbf{Communicated variables:} $\mc{Q}_i = \left(x_i^{(k)}, d_i^{(k+1)}\right)$
        \;
        \textbf{Parameters:} $\mc{R}_i = \left(\alpha, \epsilon, K, \Bar{w}_i \right)$\;
        
        \doparallel ($\forall i \in \mc{V}$){stopping criterion is not satisfied} {
            \begingroup\abovedisplayskip=0pt
    	    \begin{flalign*}
        	    D_i^{(k+1)} &= \alpha \nabla^2 f_i(x_i^{(k)}) + 2\Bar{w}_{ii} I &
    	    \end{flalign*}
            \endgroup
    	    Communicate $x_i^{(k)}$ to all $j \in \mc{N}_i$
    	    \begin{flalign*}
        	    g_i^{(k+1)} &= \alpha \nabla f_i(x_{i}^{(k)}) + \sum_{j \in \mc{N}_i \cup \{i\}} \Bar{w}_{ij} x_j^{(k)} &\\
        	    d_i^{(0)} &= -\left(D_i^{(k+1)}\right)^{-1} g_i^{(k + 1)} &
    	    \end{flalign*}
    	    \For{$p = 0$ \KwTo $K - 1$} {
        	    Communicate $d_i^{(p)}$ to all $j \in \mc{N}_i$
        	    \begin{flalign*}
            	    d_i^{(p + 1)} &= \left(D_i^{(k+1)}\right)^{-1} \Biggl[\bar{w}_{ii} d_i^{(p)} - g_i^{(k+1)} &\\
            	    & \hspace{8em} - \sum_{j \in \mc{N}_i \cup \{i\}} \Bar{w}_{ij} d_j^{(p)} \Biggr] &
        	    \end{flalign*}
    	    }
    	   \begingroup\abovedisplayskip=0pt
    	    \begin{flalign*}
    	        x_i^{(k + 1)}  &= x_i^{(k)} + \epsilon \: d_i^{(K)} &
    	    \end{flalign*}
    	    \endgroup
    	    
    	    $k \gets k + 1$
        }
    \end{algorithm}

    However, the Newton descent step requires computing the inverse of the joint problem's Hessian which cannot be directly computed in a distributed manner as its inverse is dense. To allow for distributed computation of the Hessian inverse, NN-$K$ uses the first $K$ terms of the Taylor series expansion ${(I - X)^{-1} = \sum_{j=0}^{\infty}X^{j}}$ to compute the approximate Hessian inverse, as introduced in \cite{zargham2013accelerated}. Approximation of the Hessian inverse comes at an additional communication cost, and requires an additional $K$ communication rounds per update of the primal variable. Algorithm \ref{alg:NetNewtonPa} summarizes the update procedures in the NN-$K$ method in which $\epsilon$ denotes the local step-size for the Newton's step. Selection of the step-size parameter does not require any coordination between robots. As presented in Algorithm \ref{alg:NetNewtonPa}, NN-$K$ proceeds through two sets of update equations: an outer set of updates that initializes the Hessian approximation and computes the decision variable update and an inner Hessian approximation update; a communication round precedes the execution of either set of update equations.  Increasing $K$, the number of intermediary communication rounds, improves the accuracy of the approximated Hessian inverse at the cost of increasing the communication cost per primal variable update. 
    
    A follow-up work optimizes a quadratic approximation of the augmented Lagrangian of the general distributed optimization problem \eqref{eq:general_problem} where the primal variable update involves computing a $P$-approximate Hessian inverse to perform a Newton descent step, and the dual variable update uses gradient ascent \cite{Mokhtari2016}. The resulting algorithm Exact Second-Order Method (ESOM) provides a faster convergence rate than NN-$K$ at the cost of one additional round of communication for the dual ascent step. Notably, replacing the augmented Lagrangian in the ESOM formulation with its linear approximation results in the EXTRA update equations, showing the relationship between both approaches.
    
    In some cases, computation of the Hessian is impossible because second-order information is not available {\color{black} or intractable due to the large dimensions of the problem}. Quasi-Newton methods like the Broyden-Flectcher-Goldman-Shanno (BFGS) algorithm approximate the Hessian when it cannot be directly computed, e.g., \cite{shorinwa2024distributedquasinewton}. The distributed BFGS (D-BFGS) algorithm \cite{Eisen2017} replaces the second-order information in the primal update in ESOM with a BFGS approximation (i.e., replaces $D_i^{(k)}$ in a call to the Hessian approximation equations in Algorithm \ref{alg:NetNewtonPa} with an approximation), and results in essentially a ``doubly" approximate Hessian inverse. In \cite{Eisen2019} the D-BFGS method is extended so that the dual update also uses a distributed Quasi-Newton update scheme, rather than gradient ascent. The resulting primal-dual Quasi-Newton method requires two consecutive iterative rounds of communication doubling the communication overhead per primal variable update compared to its predecessors (NN-$K$, ESOM, and D-BFGS). However, the resulting algorithm is shown by the authors to still converge faster in terms of required communication. In addition, asynchronous variants of the approximate Newton methods have been developed \cite{mansoori2019fast}.

    \subsection{Convex Surrogate Methods}  
    While the approximate Newton methods in \cite{Mokhtari2016, Eisen2017, Eisen2019} optimize a quadratic approximation of the augmented Lagrangian of \eqref{eq:penalty_distributed_problem}, other distributed methods allow for more general and direct convex approximations of the distributed optimization problem. These convex approximations generally require the gradient of the joint objective function which is inaccessible to any single robot. In the NEXT family of algorithms \cite{di2016next} dynamic consensus is used to allow each robot to approximate the global gradient, and that gradient is then used to compute a convex approximation of the joint objective function locally. A variety of surrogate functions, $U(\cdot)$, are proposed including linear, quadratic, and block-convex functions, which allows for greater flexibility in  tailoring the algorithm to individual applications. Using its surrogate of the joint objective function, each robot updates its local variables iteratively by solving its surrogate for the problem, and then taking a weighted combination of the resulting solution with the solutions of its neighbors. To ensure convergence, NEXT algorithms require a series of decreasing step-sizes, resulting in generally slower convergence rates as well as additional hyperparameter tuning.

    \begin{algorithm}[t]
    	\caption{NEXT}\label{alg:NEXT}
    	\textbf{Initialization:} $k \gets 0$, ${x_{i}^{(0)} \in \mathbb{R}^{n}}$, ${y_{i}^{(0)} = \nabla f_{i}(x_{i}^{(0)})}$, {\phantom{Initialization:}} ${\ti{\pi}_i^{(k+1)} = N y_{i}^{(0)} - \nabla f_{i}(x_{i}^{(0)})}$
    	\;
        \textbf{Internal variables:} $\mc{P}_i = \left(x_i^{(k)}, \tilde{x}_i^{(k)}, \ti{\pi}_i^{(k)} \right)$
        \;
        \textbf{Communicated variables:} $\mc{Q}_i^{(k)} = \left(z_i^{(k)}, y_i^{(k)} \right)$
        \;
        \textbf{Parameters:} $\mc{R}_i^{(k)} = \left(\alpha^{(k)}, w_i, U(\cdot),  \mathcal{X}_{i} \right)$\;
        
        \doparallel ($\forall i \in \mc{V}$){stopping criterion is not satisfied} {
            \begingroup\abovedisplayskip=0pt
            \begin{flalign*}
                \ti{x}_i^{(k)} &= \argmin_{x \in \mathcal{X}_{i}}\ U \left( x; x_i^{(k)}, \ti{\pi}_i^{(k)} \right) &\\
                z_i^{(k)} &= x_i^{(k)} + \alpha^{(k)} \left(\ti{x}_i^{(k)} - x_i^{(k)} \right) &
            \end{flalign*}
            \endgroup
    	    Communicate $\mc{Q}_i^{(k)}$ to all $j \in \mc{N}_i$\;
    	    Receive $\mc{Q}_j^{(k)}$ from all $j \in \mc{N}_i$
    	    \begin{flalign*}
                x_i^{(k+1)} &= \sum_{j \in \mc{N}_{i} \cup \{i\}} w_{ij} z_j^{(k)} &\\
                y_i^{(k+1)} &= \sum_{j \in \mc{N}_{i} \cup \{i\}} w_{ij} y_j^{(k)} &\\
                & \qquad + \left[ \nabla f_i(x_i^{(k + 1)}) -  \nabla f_i(x_i^{(k)}) \right] &\\
                \ti{\pi}_i^{(k+1)} &= N \cdot y_i^{(k + 1)} - \nabla f_i(x_i^{(k+1)}) &
            \end{flalign*}
    	    
    	    $k \gets k + 1$
        }
    \end{algorithm}
    
   The SONATA \cite{sun2017distributed} algorithm extends the surrogate function principles of NEXT, and proposes a variety of non-doubly-stochastic weighting schemes that can be used to perform gradient averaging similar to the push-sum protocols. The authors of SONATA also show that several configurations of the algorithm result in already proposed distributed optimization algorithms including Aug-DGM \cite{xu2015augmented}, Push-DIG \cite{nedic2017digging}, and ADD-OPT \cite{xi2017add}.

	\section{Alternating direction method of multipliers}\label{sec:ADMM}

    Considering the optimization problem in \eqref{eq:general_problem_with_agreement} with only agreement constraints, we have
    \begin{align}
        \label{eq:agreement_problem}
        \min_{\{x_i \in \mathbb{R}^{n},\  \forall i \in \mathcal{V}\}} &\sum_{i \in \mathcal{V}} f_i(x_i)\\
        \text{subject to } & x_i = x_j  \qquad\forall(i, j) \in \mathcal{E}.
    \end{align}
    The \textit{method of multipliers} solves this problem by alternating between minimizing the augmented Lagrangian of the optimization problem with respect to the primal variables $x_1, \dots, x_n$ (the ``primal update") and taking a gradient step to maximize the augmented Lagrangian with respect to the dual (the ``dual update''). The augmented Lagrangian of \eqref{eq:agreement_problem} is given by
	\begin{equation}
	    \label{eq:augmented_lagrangian}
	    \begin{aligned}
        	\mathcal{L}_{a}(\mathbf{x},q) &=  \sum_{i = 1}^{N} f_{i}(x_{i}) \\
        	& \enspace + \sum_{i = 1}^{N} \sum_{j \in \mathcal{N}_{i}} \left(q_{i,j}^{\top}(x_{i} - x_{j}) +  \frac{\rho}{2}  \norm{x_{i} - x_{j}}_{2}^{2}\right),
    	\end{aligned}
	\end{equation}
	where $q_{i,j}$ represents a dual variable for the consensus constraints between robots $i$ and $j$, ${q = \left[q_{i,j}^{\top},\ \forall (i,j) \in \mathcal{E}\right]^{\top}}$, and ${\mathbf{x} = \left[x_{1}^{\top},x_{2}^{\top},\cdots,x_{N}^{\top}\right]^{\top}}$. The parameter ${\rho > 0}$ represents a penalty term on the violations of the consensus constraints.
    {\color{black} The quadratic penalty term is what distinguishes the augmented Lagrangian, and it also distinguishes the method of multipliers from dual ascent. The main benefit of using the augmented Lagrangian is that the quadratic term essentially serves as a proximal operator and helps to ensure convergence.}

    In the \textit{alternating direction method of multipliers} (ADMM), given the separability of the global objective function, the primal update is executed as successive minimizations over each primal variable (i.e., choose the minimizing $x_1$ with all other variables fixed, then choose the minimizing $x_2$, and so on). Most ADMM-based approaches do not satisfy our definition of distributed in that either the primal updates take place sequentially rather than in parallel or the dual update requires centralized computation \cite{houska2016augmented,chatzipanagiotis2015augmented, iutzeler2013asynchronous}.  However, the \textit{consensus alternating direction method of multipliers} (C-ADMM) provides an ADMM-based optimization method that is fully distributed: the nodes alternate between updating their primal and dual variable and communicating with neighboring nodes \cite{mateos2010distributed, terelius2011decentralized}.

    In order to achieve a distributed update of the primal and dual variables, C-ADMM alters the agreement constraints between agents with an existing communication link by introducing auxiliary primal variables in \eqref{eq:general_problem_with_agreement} (instead of the constraint $x_i = x_j$, we have two constraints: $x_i = z_{ij}$ and $x_j = z_{ij}$).  Considering the optimization steps across the entire network, \mbox{C-ADMM} proceeds by optimizing the original primal variables, then the auxiliary primal variables, and then the dual variables, as in the original formulation of ADMM.  We can perform minimization with respect to the primal variables and gradient ascent with respect to the dual variables on an augmented Lagrangian that is fully distributed among the robots. {\color{black}Further, we note that although ADMM is typically applied to equality-constrained problems, the method can be extended to inequality-constrained problems quite easily. In particular, we note that inequality-constrained problems can be expressed as equality-constrained problems using indicator functions. With this approach, corresponding update procedures for constrained optimization problems can be derived using ADMM.}
    
    Algorithm \ref{alg:C-ADMM} summarizes the update procedures for the local primal and dual variables of each agent in constrained optimization problems, where $y_{i}$ represents the dual variable that enforces agreement between robot $i$ and its neighbors. We have incorporated the solution of the auxiliary primal variable update into the update procedure for $x_{i}^{(k+1)}$, noting that the auxiliary primal variable update can be performed implicitly ($z_{ij}^* = \frac{1}{2}\left(x_{i} + x_j\right)$). The parameter $\rho$ that weights the quadratic terms in $\mathcal{L}_a$ is also the step-size in the gradient ascent of the dual variable. We note that the update procedure for $x_{i}^{(k+1)}$ requires solving an optimization problem which might be computationally intensive for certain objective functions. To simplify the update complexity, the optimization can be solved inexactly using a linear approximation of the objective function such as \cite{Ling2015, chang2014multi, farina2019distributed} or a quadratic approximation using the Hessian such as DQM \cite{MokhtariDQM2016}. The convergence rate of ADMM methods depends on the value of the penalty parameter $\rho$. Several works discuss effective strategies for optimally selecting $\rho$ \cite{teixeira2015admm}. In general, convergence of C-ADMM and its variants is only guaranteed when the dual variables sum to zero, a condition that could be challenging to satisfy in problems with unreliable communication networks. Other distributed ADMM variants which do not require this condition have been developed \cite{meng2015proximal, makhdoumi2017convergence}. However, these methods incur a greater communication overhead to provide robustness in these problems. Gradient tracking algorithms are related to C-ADMM, when the minimization problem in the primal update procedure is solved using a single gradient decent update.

    \begin{algorithm}[t]
    	\caption{C-ADMM} \label{alg:C-ADMM}
    	\textbf{Initialization:} $k \gets 0$, ${x_{i}^{(0)} \in \mathbb{R}^{n}}$, ${y_{i}^{(0)} = 0}$
    	\;
        \textbf{Internal variables:} $\mc{P}_i^{(k)} = y_i^{(k)}$
        \;
        \textbf{Communicated variables:} $\mc{Q}_i^{(k)} = x_i^{(k)}$
        \;
        \textbf{Parameters:} $\mc{R}_i^{(k)} = \rho$\;
        
        \doparallel ($\forall i \in \mc{V}$){stopping criterion is not satisfied} {
            \begingroup\abovedisplayskip=0pt
            \begin{flalign*} 
                &x_i^{(k+1)} = \argmin_{x_i \in \mathcal{X}_{i}} \Bigg\{f_i(x_i) + x_i^\top y_i^{(k)} \cdots \\
                & \hspace{6em} + \rho \sum_{j \in \mc{N}_i} \left\Vert x_i - \frac{1}{2}\left(x_i^{(k)} + x_j^{(k)}\right)\right\Vert^2_{2} \Bigg\} &
            \end{flalign*}
            \endgroup
    	    Communicate $\mc{Q}_i^{(k)}$ to all $j \in \mc{N}_i$\;
    	    Receive $\mc{Q}_j^{(k)}$ from all $j \in \mc{N}_i$
    	    \begin{flalign*}
                y_i^{(k+1)} &= y_i^{(k)} + \rho\sum_{j \in \mc{N}_i} \left(x_i^{(k + 1)} - x_j^{(k + 1)}\right) &
            \end{flalign*}
    	    
    	    $k \gets k + 1$
        }
    \end{algorithm}
    
    C-ADMM, as presented in Algorithm \ref{alg:C-ADMM}, requires each robot to optimize over a local copy of the global decision variable $x$. However, many robotic problems have a fundamental structure that makes maintaining global knowledge at every individual robot unnecessary: each robot's data relate only to a subset of the global optimization variables, and each agent only requires a subset of the optimization variable for its role.  For instance, in distributed SLAM, a memory-efficient solution would require a robot to optimize only over its local map and communicate with other robots only messages of shared interest. Other examples arise in distributed environmental monitoring by multiple robots \cite{elwin2019distributed}. The SOVA method \cite{ola2020SOVA} leverages the separability of the optimization variable to achieve orders of magnitude improvement in convergence rates, computation, and communication complexity over C-ADMM methods. 
    {\color{black} The general approach of SOVA can also be found in partitioning-based methods such as in \cite{erseghe2012distributed, bastianello2018partition, todescato2020partition}, which also accomodate asynchronous or lossy communication. Like SOVA, these methods exploit the partitioning of the state variables, in that robot $i$ need not estimate the states that are not relevant to its local objective function.}
    
    In SOVA, each agent only optimizes over variables relevant to its data or role, enabling robotic applications in which agents have minimal access to computation and communication resources. SOVA introduces consistency constraints between each agent's local optimization variable and its neighbors, mapping the elements of the local optimization variables, given by
    \begin{align*}
        \Phi_{ij}x_{i} = \Phi_{ji}x_{j} \quad \forall j \in \mc{N}_{i},\ \forall i \in \mc{V}
    \end{align*}
    where $\Phi_{ij}$ and $\Phi_{ji}$ map elements of $x_{i}$ and $x_{j}$ to a common space. C-ADMM represents a special case of SOVA where $\Phi_{ij}$ is always the identity matrix. The update procedures for each agent reduce to
    the equations given in Algorithm \ref{alg:SOVA}.

    \begin{algorithm}[t]
    	\caption{SOVA} \label{alg:SOVA}
    	\textbf{Initialization:} $k \gets 0$, ${x_{i}^{(0)} \in \mathbb{R}^{n_{i}}}$, ${y_{i}^{(0)} = 0}$
    	\;
        \textbf{Internal variables:} $\mc{P}_i^{(k)} = y_i^{(k)}$
        \;
        \textbf{Communicated variables:} $\mc{Q}_i^{(k)} = x_i^{(k)}$
        \;
        \textbf{Parameters:} $\mc{R}_i^{(k)} = \rho$\;
        
        \doparallel ($\forall i \in \mc{V}$){stopping criterion is not satisfied} {
            \begingroup\abovedisplayskip=0pt
            \begin{flalign*} 
                &x_i^{(k+1)} = \argmin_{x_i \in \mathcal{X}_{i}} \Bigg\{f_i(x_i) + x_i^\top y_i^{(k)} \cdots \\
                & \hspace{2em} + \rho \sum_{j \in \mc{N}_i} \left\Vert \Phi_{ij} x_i - \frac{1}{2}\left(\Phi_{ij}x_i^{(k)} + \Phi_{ji}x_j^{(k)}\right)\right\Vert^2_{2} \Bigg\} &
            \end{flalign*}
            \endgroup
    	    Communicate $\mc{Q}_i^{(k)}$ to all $j \in \mc{N}_i$\;
    	    Receive $\mc{Q}_j^{(k)}$ from all $j \in \mc{N}_i$
    	    \begin{flalign*}
                y_i^{(k+1)} &= y_i^{(k)} + \rho\sum_{j \in \mc{N}_i} \Phi_{ij}^\top \left(\Phi_{ij}x_i^{(k)} - \Phi_{ji}x_j^{(k)}\right) &
            \end{flalign*}
    	    
    	    $k \gets k + 1$
        }
    \end{algorithm}

    \section{Distributed Optimization in Robotics and Related Applications}
    \label{sec:applications_in_literature}
    In this section, we discuss some existing applications of distributed optimization to robotics problems. To simplify the presentation, we highlight a number of these applications in the following notable problems in robotics: synchronization, localization, mapping, and target tracking; online and deep learning problems; and task assignment, planning, and control. We refer the reader to the first paper in this two-part series \cite{shorinwa_distributed_2023} for a case study on multi-drone target tracking, which compares solutions using several different distributed optimization algorithms. 
    
    \subsection{Synchronization, Localization, Mapping, and Tracking}
    Distributed optimization algorithms have found notable applications in robot localization from relative measurements \cite{dang2016decentralized, alwan2015distributed}, including in networks with asynchronous communication \cite{todescato2015distributed}. More generally, distributed first-order algorithms have been applied to optimization problems on manifolds, including $SE(3)$ localization \cite{tron2009distributed, tron2011distributed, tron2012distributed, tron2014distributed}, synchronization problems \cite{sarlette2009consensus}, and formation control in $SO(3)$ \cite{oh2013formation, oh2018distributed}. In pose graph optimization, distributed optimization has been employed through majorization-minimization schemes, which minimize an upper-bound of the objective function \cite{fan2020majorization}; using gradient descent on Riemannian manifolds \cite{tian2020asynchronous, knuth2013collaborative}; and block-coordinate descent \cite{tian2019block}. Other pose graph optimization methods have utilized distributed sequential programming algorithms using a quadratic approximation model of the non-convex objective function with Gauss-Seidel updates to enable distributed local computations among the robots \cite{choudhary2017distributed}. Further, ADMM has been employed in bundle adjustment and pose graph optimization problems, which involve the recovery of the 3D positions and orientations of a map and camera \cite{zhang2017distributed, eriksson2016consensus, choudhary2015exactly}. However, many of these algorithms require a central node for the dual variable updates, making them semi-distributed. Nonetheless, a few fully-distributed ADMM-based algorithms exist for bundle adjustment and cooperative localization problems \cite{natesan2017distributed, kumar2016asynchronous}. Other applications of distributed optimization arise in target tracking \cite{ola2020targettracking, shorinwa2023distributedtargettracking}, signal estimation \cite{mateos2010distributed}, and parameter estimation in global navigation satellite systems \cite{khodabandeh2019distributed}.
    
    \subsection{Online and Deep Learning Problems}
     Distributed optimization has also been applied in online, dynamic problems. In these problems, each robot gains knowledge of its time-varying objective function in an online fashion, after taking an action or decision. A number of distributed first-order algorithms have been designed for these problems \cite{lu2019online, shahrampour2017distributed, zhang2019distributed}. Similarly, DDA has been adapted for online scenarios with both static communication graphs \cite{hosseini2013online, shahrampour2013exponentially} and time-varying communication topology \cite{hosseini2016online}, \cite{lee2017stochastic}. The push-sum variant of dual averaging has also been used for distributed training of deep-learning algorithms, and has been shown to be useful in avoiding pitfalls of other synchronous distributed training frameworks, which face notable challenges in problems with communication deadlocks \cite{tsianos2012application}. {\color{black} Many of these algorithms emphasize parallelization.}
     
     In addition, distributed sequential convex programming algorithms have been developed for a number of learning problems where data is distributed, including semi-supervised support vector machines \cite{scardapane2016distributed}, neural network training \cite{scardapane2017framework}, and clustering \cite{altilio2019distributed}. Moreover, ADMM has been applied to online problems, such as estimation and surveillance problems involving wireless sensor networks \cite{ling2013decentralized, Xu2015}.  ADMM has also be applied to distributed deep learning in robot networks in \cite{yu2022dinno}. 
    
    \subsection{Task Assignment, Planning, and Control}
    Distributed optimization has been applied to task assignment problems, posed as optimization problems. Some works \cite{montijano2014efficient} employ distributed optimization using a distributed simplex method \cite{burger2012distributed} to obtain an optimal assignment of the robots to a desired target formation.  Other works employ C-ADMM for distributed task assignment \cite{ola2019task,olaTRO2023task}. Further applications of distributed optimization arise in motion planning \cite{bento2013message}, trajectory tracking problems involving teams of robots using non-linear model predictive control \cite{ferranti2018coordination, shorinwa2023distributedmodelpredictive}, and collaborative manipulation \cite{ola2020collab,shorinwa2021distributed}, which employ fully-distributed variants of ADMM.  {\color{black} One feature common to these problems is that the joint decision variables, which consists of control inputs or action variables concatenated over all the robots, can often be partitioned so that each robot only needs to consider its own actions, as in \cite{ola2020SOVA,erseghe2012distributed, bastianello2018partition, todescato2020partition}.  This can lead to significantly faster convergence compared methods in which each agent has a complete copy of the joint decision variables, as discussed as the end of Sec.~\ref{sec:ADMM} above.}

{\color{black}\section{Research Opportunities in Distributed Optimization for Multi-Robot Systems}}
\label{sec:open_problems}

{\color{black} In this section, we highlight challenges in the application of existing distributed optimization algorithms to multi-robot problems, each of which represents a promising direction for future research.}

\subsection{Non-Convex and Constrained Robotics Problems}
Distributed optimization methods have primarily focused on solving unconstrained convex optimization problems, which constitute a limited subset of robotics problems. Many robotics problems involve non-convex objectives or constraints. For example, problems in multi-robot motion planning, SLAM, learning, distributed manipulation, and target tracking are often non-convex and/or constrained. 

Both DFO methods and C-ADMM methods can be modified for non-convex and constrained problems; however, few examples of practical algorithms or rigorous analyses of performance for such modified algorithms exist in the literature.  %
One way to implement C-ADMM for non-convex problems is to solve each primal update step as a non-convex optimization (e.g., through a quasi-Newton method, or interior point method).  Another option is to perform successive quadratic approximations in an outer loop, and use C-ADMM to solve each resulting quadratic problem in an inner loop.  The trade-off between these two options has not yet been explored in the literature, especially in the context of non-convex problems in robotics.
    
\subsection{Bandwidth-Constrained, Lossy, or Dynamic Communication}
	In many robotics problems, each robot exchanges information with its neighbors over a communication network with a limited communication bandwidth, which effectively limits the size of the message packets that can be transmitted between robots. Moreover, in practical situations, the communication links between robots sometimes fail, resulting in packet losses. However, many distributed optimization methods do not consider communication between agents as an expensive, unreliable resource, given that many of these methods were developed for problems with reliable communication infrastructure (e.g., multi-core computing, or computing in a hard-wired cluster). Information quantization has been extensively employed in many disciplines to allow for efficient exchange of information over bandwidth-constrained networks. Quantization involves encoding the data to be transmitted into a format which utilizes a fewer number of bits, often resulting in lower precision. Transmission of the encoded data incurs a lower communication overhead, enabling each robot to communicate with its neighbors within the bandwidth constraints. A few distributed optimization algorithms have been designed for these problems, including quantized distributed first-order algorithms. Some of these algorithms assume that all robots can communicate with a central node \cite{alimisis2021communication, yu2019double}, making them unsuitable for a variety of robotics of problems, while others do not make this assumption \cite{pu2016quantization, reisizadeh2019exact, lee2018finite, li2017event}. In addition, quantized distributed variants of ADMM also exist \cite{zhu2016quantized, elgabli2020q, zhu2015distributed}.
 
 Generally, quantization introduces error between each robot's solution and the optimal solution. However, in some of these algorithms, the quantization error decays during the execution of the algorithms under certain assumptions on the quantizer and the quantization interval \cite{pu2016quantization, reisizadeh2019exact}. However, quantization in distributed optimization algorithms generally results in slower convergence rates, which poses a challenge in robotics problems where a solution is required rapidly, such as model predictive control problems, highlighting the need for the development of more effective algorithms. Further, only a few distributed optimization algorithms consider problems with lossy communication networks \cite{bastianello2018distributed, bastianello2020asynchronous, bof2018multiagent}.

 {\color{black} In many practical situations, the communication network between robots changes as robots move, giving rise to a time-varying communication graph. While many distributed first-order optimization algorithms \cite{nedic2017digging} and some distributed sequential programming algorithms \cite{di2016next, sun2017distributed} tolerate dynamic communication networks under the condition of bounded connectivity ,
 in general, distributed ADMM algorithms are not amenable to problems with dynamic communication networks.  This is an interesting avenue for future research.}

\subsection{Limited Computation Resources}
Another valuable direction for future research is in developing algorithms specifically for computationally limited robotic platforms, in which the timeliness of the solution is as important as the solution quality \cite{trenkwalder2019computational, lahijanian2018resource}.  In general, many distributed optimization methods involve computationally challenging procedures that require significant computational power, especially distributed methods for constrained problems \cite{Ling2015, chang2014multi, farina2019distributed}. These methods ignore the significance of computation time, assuming that agents have access to significant computational power. These assumptions often do not hold in robotics problems. Typically, robotics problems unfold over successive time periods with an associated optimization phase at each step of the problem. As such, agents must compute their solutions fast enough to proceed with computing a reasonable solution for the next problem which requires efficient distributed optimization methods. Developing such algorithms specifically for multi-robot systems is an interesting topic for future work.

{\color{black}
\subsection{Coordination and synchronization}
Many distributed optimization algorithms implicitly assume coordination in several aspects of implementation. First, while most algorithms accommodate an arbitrary initialization of the initial solution of each robot (at least in convex problems), they often place stringent requirements on the initialization of the algorithms' parameters. For instance, DFO methods assume a common step size across all robots and in some cases a scheduled decrease in that step size \cite{nedic2009, nedic2017achieving, shi2015extra}.
Similarly, distributed first-order algorithms and distributed sequential convex programming algorithms require the specification of a stochastic matrix, which must be compatible with the underlying communication network. However, generating doubly-stochastic matrices for directed communication networks is nontrivial if each robot does not know the global network topology \cite{gharesifard2012distributed}.
ADMM and its distributed variants require the selection of a common penalty parameter $\rho$.

Second, some distributed first-order, distributed sequential programming, and distributed ADMM algorithms require synchronous execution (see Definition \ref{def:sync}). If robots have variable computation times and a synchronous distributed optimization algorithm is being used, one solution is to implement a distributed barrier scheme where each robot waits until all of its neighbors have computed and communicated their most recent update before proceeding. However, barrier schemes can lead to significantly increased time to convergence as some robots idle while waiting for their neighbors. To address this issue, a number of asynchronous distributed optimization algorithms have been developed \cite{lian2018asynchronous, zheng2017asynchronous, Eisen2017, mansoori2019fast, kumar2016asynchronous}, which allow each robot to perform its local updates asynchronously, eliminating the need for synchronization. These asynchronous variants are guaranteed to converge to an optimal solution, provided that an integer ${T \in \mathbb{Z}}$ exists such that each robot performs at least one iteration of the algorithm over $T$ time-steps.

}

\subsection{Hardware Implementation}
 {\color{black} Finally, we believe there is a gap between the analysis in the distributed optimization literature and the applicability of these distributed optimization algorithms to hardware implementations \cite{nedic2018network, nedic2018distributedcollection, yang2019survey}. The suitability  of algorithms to run efficiently and robustly on robots has still not be thoroughly proven.  We provide empirical results of a hardware implementation of C-ADMM over XBee radios in the first paper in this series \cite{shorinwa_distributed_2023}. While this survey considers adapting existing distributed optimization algorithms for robotic implementations, it could also be useful to consider the co-design of general purpose distributed optimization algorithms with practical hardware setups.}

\section{Conclusion}
\label{sec:conclusion}
Despite the amenability of many robotics problems to distributed optimization, few applications of distributed optimization to multi-robot problems exist. In this work, we have categorized distributed optimization methods into three broad classes---distributed first-order methods, distributed sequential convex programming methods, and the alternating direction method of multipliers (ADMM)---highlighting the distinct mathematical techniques employed by these algorithms. 
Further, we have identified a number of important open challenges in distributed optimization for robotics, which could be interesting areas for future research. In general, the
opportunities for research in distributed optimization for multi-robot systems are plentiful. Distributed optimization provides
an appealing unifying framework from which to synthesize
solutions for a large variety of problems in multi-robot systems.

	\bibliographystyle{IEEEtran}
	\bibliography{references}

\end{document}